\documentclass{article} % For LaTeX2e
\usepackage[table]{xcolor}
\usepackage{iclr2024_conference,times}

\usepackage{amsmath,amsfonts,bm}

\def\eqref#1{equation~\ref{#1}}
\def\1{\bm{1}}

\DeclareMathAlphabet{\mathsfit}{\encodingdefault}{\sfdefault}{m}{sl}
\SetMathAlphabet{\mathsfit}{bold}{\encodingdefault}{\sfdefault}{bx}{n}

\usepackage{hyperref}
\hypersetup{
    colorlinks=true,
    linkcolor=orange,
    filecolor=magenta,      
    urlcolor=orange,
    citecolor=orange,
}
\usepackage[capitalise, nameinlink]{cleveref}
\usepackage{url}
\usepackage{graphicx}
\usepackage{booktabs}
\usepackage{multirow}
\usepackage{bbm}
\usepackage{titlesec}
\usepackage{pifont} 

\titlespacing*{\section}
{0pt}{1.5ex plus 1ex minus .2ex}{1ex plus .2ex}

\titlespacing*{\subsection}
{0pt}{1.25ex plus 1ex minus .2ex}{0.75ex plus .2ex}

\titlespacing*{\subsubsection}
{0pt}{1ex plus 0.5ex minus .2ex}{0.5ex plus .2ex}

\title{What's the Move?\\Hybrid Imitation Learning via Salient Points}

\author{Priya Sundaresan$^{*1}$, Hengyuan Hu$^{*1}$, Quan Vuong$^{2}$, Jeannette Bohg$^{1}$, Dorsa Sadigh$^{1}$ \\
$^*$Equal contribution \\
$^1$Stanford University, $^2$Physical Intelligence \\
}

\makeatletter
\def\thanks#1{\protected@xdef\@thanks{\@thanks\protect\footnotetext{#1}}}
\makeatother
\thanks{Correspondence to \texttt{\{priyasun, hengyuan\}@stanford.edu}}

\newcommand{\algabbr}{\textbf{\textsc{Sphinx}}}
\iclrfinalcopy % Uncomment for camera-ready version, but NOT for submission.
\begin{document}

\maketitle
\begin{abstract}
While imitation learning (IL) offers a promising framework for teaching robots various behaviors, learning complex tasks remains challenging. Existing IL policies struggle to generalize effectively across visual and spatial variations even for simple tasks. In this work, we introduce \algabbr~(Salient Point-Based Hybrid ImitatioN and eXecution), a flexible IL policy that leverages multimodal observations (point clouds and wrist images), along with a hybrid action space of low-frequency, sparse waypoints and high-frequency, dense end effector movements. Given 3D point cloud observations, \algabbr~learns to infer task-relevant points within a point cloud, or \emph{salient points}, which support spatial generalization by focusing on semantically meaningful features. These salient points serve as anchor points to predict waypoints for long-range movement, such as reaching target poses in free-space. Once near a salient point, \algabbr~learns to switch to predicting dense end-effector movements given close-up wrist images for precise phases of a task. By exploiting the strengths of different input modalities and action representations for different manipulation phases, \algabbr~tackles complex tasks in a sample-efficient, generalizable manner. Our method achieves $\mathbf{86.7\%}$  success across 4 real-world and 2 simulated tasks, outperforming the next best state-of-the-art IL baseline by $\mathbf{41.1\%}$ on average across $\mathbf{440}$ real world trials. \algabbr~additionally generalizes to novel viewpoints, visual distractors, spatial arrangements, and execution speeds with a $\mathbf{1.7\times}$ speedup over the most competitive baseline. Our website contains code for data collection and training code along with supplementary videos: \href{http://sphinx-manip.github.io}{\texttt{http://sphinx-manip.github.io}}.
\end{abstract}

\section{Introduction}
\label{sec:intro}
\begin{figure}[t]
\centering
\includegraphics[width=1.0\textwidth]{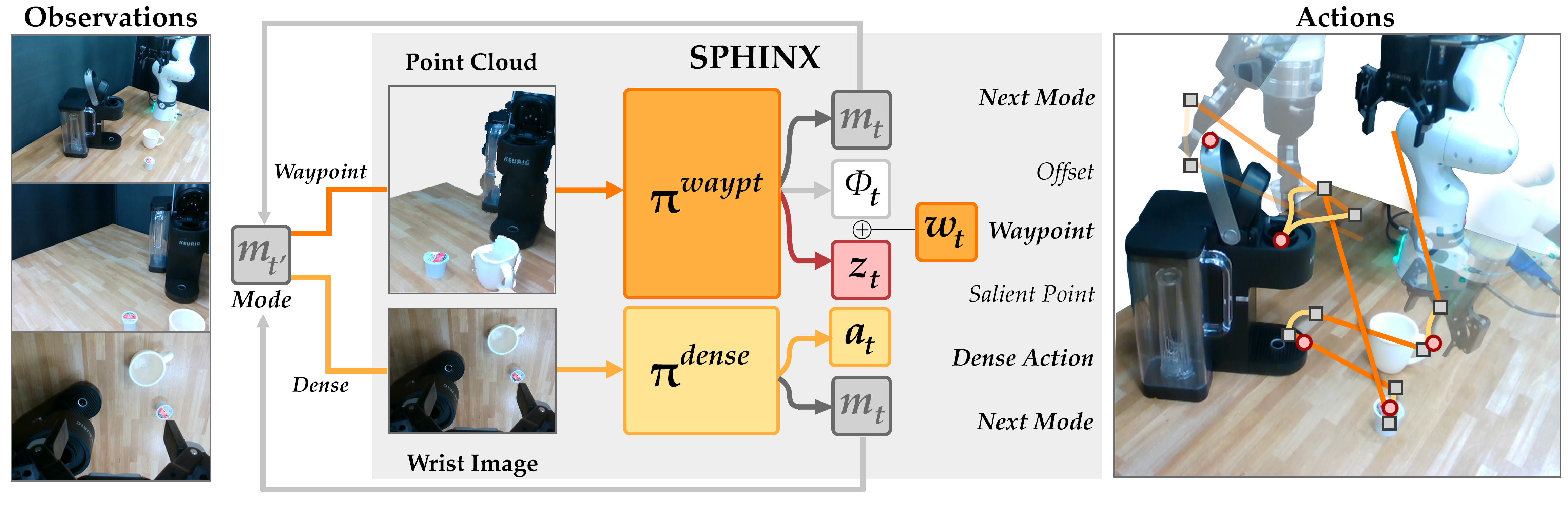}
% \vspace{-6mm}
\caption{\small \algabbr~is a hybrid IL agent which learns to switch amongst different modes ($m_t$) of execution to tackle complex tasks with visuospatial generalization. In \textcolor[rgb]{0.9, 0.4, 0.1}{\emph{waypoint}} mode, $\pi^{\mathrm{waypt}}$ takes a point cloud as input, and predicts a single \emph{waypoint} $w_t$ as an offset ($\phi_t$) to a task-relevant \emph{salient point} $z_t$ (i.e. mug handle, coffee pod, etc. denoted). After reaching a waypoint via a controller, the policy uses learned switching to a dense policy $\pi^{\mathrm{dense}}$, which takes wrist-camera images as input and outputs \textcolor[rgb]{0.9, 0.6, 0.1}{\emph{dense}} actions ($a_t$) for precise manipulation around a salient point. On the right, the policy interleaves both modes of execution to complete a long-horizon coffee-making task guided by salient points (\textcolor{pink}{\ding{108}}) and mode switches (\textcolor{lightgray}{\ding{110}}).}
\label{fig:method-high-level}
\end{figure}

Imitation learning (IL) of visuomotor policies is a widely used framework for teaching robots manipulation tasks given demonstrations collected by humans~\citep{schaal1996learning}. 
While prior works have shown that IL policies can learn a range of behaviors with sufficient data, from simple object pick-and-place to more complex tasks, they typically succeed only in highly controlled settings with low variation. Generalizing to realistic visual and spatial variations remains a significant challenge. Consider teaching a robot to make a cup of coffee in the morning, which demands precision, long-horizon reasoning, and tolerance to environment variations. The robot must first carefully grasp a mug handle, position it under the machine, insert a pod into a narrow slot, close the lid, and press a button -- all with very little margin for error (\cref{fig:method-high-level}). Even after mastering this sequence, the policy might struggle with \emph{spatial} changes like moving the machine, or \emph{visual} changes such as new coffee pods, spilled grounds, a different camera angle, or varying lighting conditions. This underscores the need for IL policies that can learn complex tasks from a limited number of demonstrations while effectively generalizing to natural and expected variations in the real-world. 

Conventional IL policies often struggle with both performance and generalization, largely due to limitations in their input and output representations. First, they tend to rely heavily on visual inputs like RGB images, treating irrelevant details like the background, lighting, or viewpoint the same as task-relevant information. This can cause a policy to memorize specific scenes, making it brittle to \emph{visual} variations~\citep{zhao2023learning, chi2023diffusion}. Second, these policies usually predict actions for the next immediate timestep, which hampers \emph{spatial} reasoning. Simple spatial movements, like reaching, are predicted through hundreds of end-effector actions, increasing the risk of veering off course. To address these limitations, recent works explore 3D scene representations, such as point clouds and voxel grids, to offer better spatial awareness, and propose predicting actions as end-effector poses (\emph{waypoints}) reachable through a controller or motion planner~\citep{goyal2023rvt, sundaresan2023kite, yang2024equibot, yang2024equivact, shridhar2023perceiver}. This can drastically shorten the action prediction horizon and enable better spatial generalization. However, these methods often lack precision, as point clouds typically lack the necessary resolution to capture small object details preserved in images. On the other hand, recent image-based IL policies attempt to remedy spatial generalization using \emph{hybrid action spaces}~\citep{belkhale2023hydra, shi2023waypoint}. Here, the policy has two potential \emph{modes} of execution: \emph{waypoints} for long-range motions like reaching, or \emph{dense} actions --- end-effector movements predicted per-timestep --- only when precision or reactivity is required. These works ultimately consider training policies with a single input modality type, and optionally hybrid actions. In reality, different phases of a task may lend themselves more favorably to different visual inputs or action modes. A policy which can effectively choose and interchange both the input modality and the underlying action representation during execution remains underexplored.

Our key insight is that by encouraging the policy to attend to \emph{salient points}—task-relevant 3D points—we enable it to choose between different input modalities as well as different action modes when appropriate, improving performance and generalization. \cref{fig:method-high-level} illustrates example salient points in the coffee-making task in red (i.e. the mug handle, pod, etc.). We introduce \algabbr: {\em Salient Point-based Hybrid ImitatioN and eXecution\/}, a hybrid IL agent which learns to switch amongst a \emph{waypoint policy} which predicts waypoint actions given point clouds, and a \emph{dense policy} which predicts dense actions given close-up wrist-camera images. Specifically, the waypoint policy manages long-range movements by first predicting salient points that narrow the search space of actions around spatially relevant features, promoting \emph{spatial} generalization. It then predicts waypoint actions relative to these points. After reaching a waypoint, \algabbr~ switches to a dense policy which takes wrist-camera images as input. This policy captures close-up object details for precise manipulation and supports \emph{visual} generalization by staying agnostic to broader scene changes. To support training \algabbr, we develop a flexible data collection interface that allows demonstrators to specify salient points and switch modes in real-time  during teleoperation. 

Empirically, we show that \algabbr~can tackle a range of precise, long-horizon manipulation tasks, including four real-world scenarios (drawer-opening, cup-stacking, coffee-making, toy train assembly) and two simulated ones. \algabbr~achieves \textbf{86.7\% success} and outperforms the next best IL baseline by \textbf{41.1\%} on average, while generalizing better to visual distractors, viewpoints, spatial arrangements, and execution speeds. We open-source our web-based data collection interface for specifying salient points and hybrid teleoperation, alongside code, supplementary material, and videos at: \href{http://sphinx-manip.github.io}{\texttt{http://sphinx-manip.github.io}}.
\vspace{-0.3cm}

\section{Related Work}
\label{sec:related}

\textbf{Imitation Learning for Robotics Control:} 
Imitation learning has long been a foundational approach in robotics for teaching robots to replicate human demonstrations~\citep{schaal1996learning, atkeson1997robot, pomerleau1988alvinn}. Robotic imitation learning policies typically take images as input and output motor commands, such as joint positions, velocities, or Cartesian end-effector poses. Recent works of that type~\citep{reuss2023goal-diffuse, chi2023diffusion, zhao2023learning} have demonstrated strong performance on tasks in controlled settings with a limited initial state distribution. However, they struggle to generalize to unseen visual or spatial variations. To address visual generalization, some works augment vision-based policies with diffusion-generated image observations~\citep{yu2023scaling, bharadhwaj2024roboagent}. While useful and complementary to our approach, these augmentations do not directly enable spatial generalization. Other works propose replacing image inputs with 3D scene representations such as point clouds and voxel grids, and outputting actions as \emph{waypoints}, 6-DoF poses reachable through motion planning~\citep{sundaresan2023kite, shridhar2023perceiver, yang2024equivact}. While this reduces the complexity of action prediction from hundreds of actions to a single pose, 3D representations such as point clouds often lack the resolution to enable precise manipulation of small objects. Other recent approaches like HYDRA~\citep{belkhale2023hydra} and AWE~\citep{shi2023waypoint} take image inputs but propose a hybrid output action space of \emph{waypoints} and \emph{dense} actions. These distinct action modes are intended for long-range and precise movements, respectively. Our method builds on these approaches by leveraging \emph{salient points} to bridge a hybrid input space of point clouds and wrist-camera images, and a hybrid output action space of waypoints and dense actions.

\textbf{Action Representations:}
Most visual imitation learning works rely on standard 6-DoF action spaces, but recent efforts explore alternatives for better spatial generalization. One approach involves predicting actions as parameterized manipulation primitives instead of low-level end-effector movements. This reduces the dimensionality of the action space and improves sample efficiency, but often requires task-specific engineering~\citep{dalal2021accelerating, sundaresan2023kite, nasiriany2022augmenting, agia2022taps}. Other methods exploit equivariance, ensuring that transformations of visual inputs (e.g., rotations or scaling) are reflected in output actions~\citep{wang2024equivariant, yang2024equibot, yang2024equivact}. However, these works often rely on limiting assumptions like access to object states via segmentation, or single-object tasks. In the grasping domain, many policies consider point clouds as inputs and an output action space defined as per-point predictions for the end-effector pose. This has proven effective for learning sample-efficient and generalizable grasping policies~\citep{grasp-point, sundermeyer2021contact}. Inspired by this, our method also parameterizes waypoint actions as offsets to salient points in a point cloud, but we critically learn a \emph{hybrid} policy which predicts \emph{both} waypoint and dense actions to tackle longer-horizon and precise tasks beyond grasping.

\textbf{Data Collection for Imitation Learning:}
Despite the advancements in action representations and spatial generalization, the success of visual imitation learning policies still hinges on the quality of teleoperated demonstrations. Human operators typically collect robot data using interfaces like virtual reality controllers~\citep{OrbikEbert2021OculusReader}, handheld devices~\citep{chi2024universal}, puppeteering setups~\citep{zhao2023learning}, or 3D mice (e.g., Spacemouse). However, these interfaces map demonstrator controls directly to robot actions on a \emph{per-timestep} basis, which presents two key limitations. First, the recorded data only captures (observation, dense action) pairs, lacking compatibility with waypoint actions or useful metadata such as salient points. Second, directly controlling long-range movements can be inefficient, noisy, and tiring for demonstrators. To address these issues, we design an interface (\cref{fig:interface}) that seamlessly integrates both waypoint and dense action modes. A custom web-based GUI supports waypoint mode, allowing demonstrators to specify salient points and waypoints with the ease of simple clicks and drags. A controller can then reach the specified waypoint automatically, removing the need for constant teleoperation from the demonstrator. Additionally, the interface is compatible with any external device for dense actions, allowing for easy switching between the computer mouse and the device, as long as it is on hand. This provides a flexible and efficient data collection process for high-quality hybrid datasets, with no post-hoc labeling required.
\section{Problem Statement}
\label{sec:preliminaries}
In standard IL, we are given a dataset $\mathcal{D}$ of $N$ trajectories of expert demonstrations $\{\tau_1, \dots, \tau_N\}$.
Each trajectory is a sequence of observation action pairs $(o_0, a_0, \dots, o_T, a_T)$. 
The goal is to learn a policy $\pi(a_t|o_t)$ that matches the expert distribution using the following loss $-\mathbb{E}_{\tau \sim \mathcal{D}}\log \pi(a_t|o_t)$.
However, this formulation can easily lead to compounding errors for long-horizon tasks where episodes may span hundreds of steps. 
In this paper, we instead consider a \emph{hybrid} imitation learning setting where the policy can either output a \textbf{dense action} $a_t \in \mathbb{R}^7$ as in the standard setting or output a \textbf{waypoint} $w_t \in \mathbb{R}^7$ that abstracts a sequence of dense actions into a single prediction for long-range movement. Both waypoint and dense actions capture the end-effector pose, but a waypoint action $w_{t'}$ specified at timestep $t'$ is translated to a sequence of $k_{t'}$ interpolated actions $\{a_{t'}, \dots, a_{t'+k_{t'}-1}\}$ by a controller $a_t = \mathcal{C}(o_t^\texttt{pose}, w_{t'})$ based on the current pose $o_t^\texttt{pose}$ and the target waypoint $w_{t'}$ specified at $t'$.
In practice, we use a simple controller that linearly interpolates between current pose and target waypoint.
Then, we can record a timestep in the dataset as $(o_t, a_t, m_t, [w_{t'}])$ where $m_t \in \{\texttt{waypt}, \texttt{dense}, \texttt{terminate} \}$ is the mode and $w_{t'}$ is the optional target waypoint only when $m_t = \texttt{waypt}$.  Each $w_{t'}$ spans the next $k_{t'}$ steps decided by the controller.
% Note the difference between $t$ and $t'$ as multiple interpolated steps correspond to one waypoint.
% The policy automatically switches between three modes,  \texttt{waypoint}, \texttt{dense}, and \texttt{terminate}. 
Our goal is thus to learn a hybrid policy $\pi(o_t)$ that first predicts a mode $p(m_t|o_t)$ and then predicts either a waypoint from $\pi^{\texttt{waypt}}(w_t|o_t)$ or a dense action from $\pi^{\texttt{dense}}(a_t|o_t)$.

\section{\algabbr: Salient Point-based Hybrid ImitatioN and eXecution}
\label{sec:method}
    
We introduce \algabbr: Salient Point-based Hybrid ImitatioN and eXecution, a framework for learning sample-efficient, generalizable imitation policies capable of handling complex, long-horizon manipulation tasks across diverse initial conditions. \algabbr~combines a high-level waypoint policy for long-range movements and a dense policy for precise manipulation (\cref{fig:method-high-level}). The waypoint policy $\pi^\texttt{waypt}$ takes point clouds  as input and classifies semantically meaningful salient points along with waypoints relative to them. This guides the robot to a suitable pose for interaction around a salient point, such as reaching for a mug handle during coffee-making. The dense policy $\pi^\texttt{dense}$ takes over only for precise actions around a salient point, like carefully inserting a coffee pod into its slot (\cref{fig:method-high-level}). Since the waypoint policy handles long-range movements, it uses point clouds to provide spatial context. The dense policy uses wrist camera images as input, capturing detailed object features for precise manipulation and enabling visual generalization to variations in the surrounding scene. Both policies also predict the next mode $m_{t+1}$ to decide which policy to use after completing the current movement. Without loss of generality, we initialize $m_0$ in waypoint mode.

To train \algabbr, we first need to collect demonstrations using the two modes and annotate salient point for each waypoint. In~\cref{sec:data-collect}, we introduce an intuitive web GUI to easily collect such demonstrations in the hybrid format and record salient points with no additional overhead. Then, we discuss how to learn $\pi^{\texttt{waypt}}(w_t|o_t)$ and $\pi^{\texttt{dense}}(a_t|o_t)$ in~\cref{sec:method-waypoint} and~\cref{sec:method-dense} respectively.

% \subsection{Problem Setup}
% \begin{itemize}
%     \item Our goal is to learn an imitation learning policy $\pi_{\theta}: \mathcal{O} \rightarrow \mathcal{A}$ mapping observations to actions, from a dataset $\mathcal{D} = \{\tau_1, \hdots, \tau_N\}$ of $N$ demonstrations
%     \item We assume access to observations $o_t = (s_t, I_t, D_t)$ where $s_t$ denotes the robot proprioceptive state (pose of the end-effector, gripper width), and $I_t$ and $D_t$ are RGB and depth images from which we can reconstruct colorized point clouds $P_t \in \mathbb{R}^{D \times 6}$ 
%     \item The action space $\mathcal{A}$ consists of robot commands (positional, velocity, joint, torque, etc.) specified as absolute poses or deltas. 
% \end{itemize}

\begin{figure}[!htbp]
\centering
\includegraphics[width=1.0\textwidth]{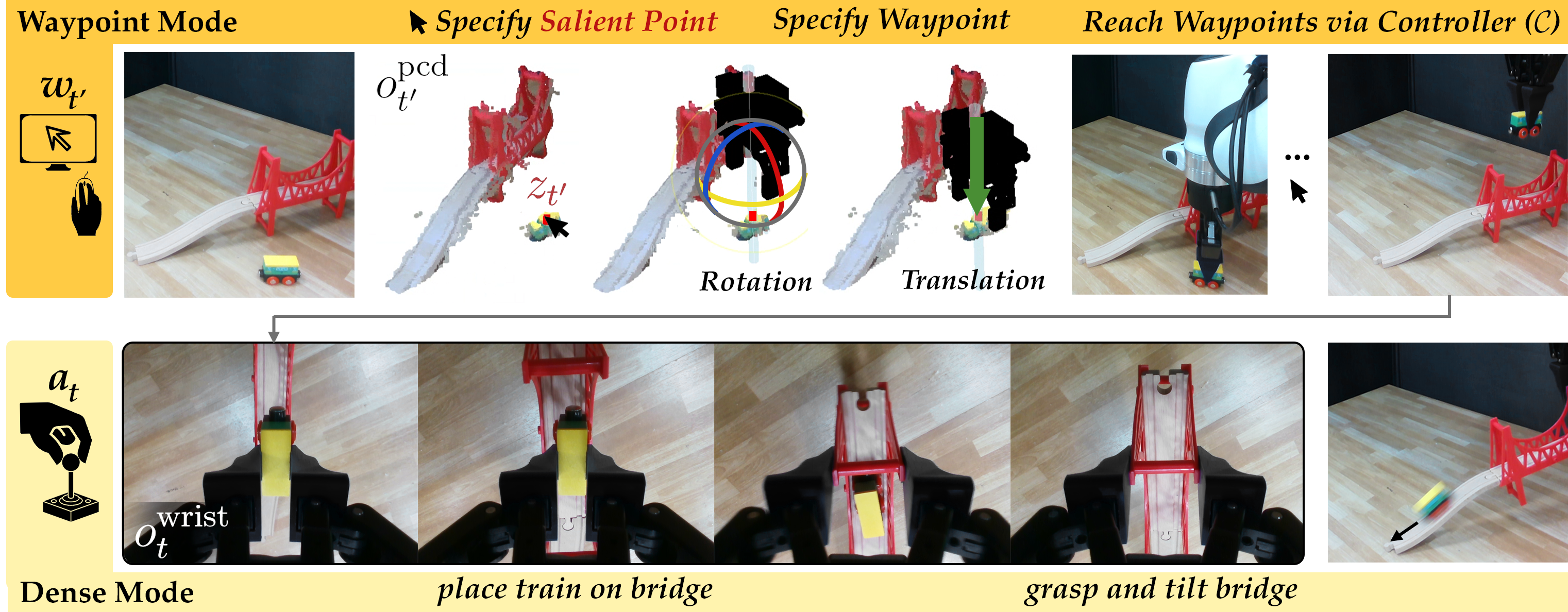}
\caption{\small \textbf{Data Collection Interface}: The demonstrator visualizes a point cloud $o^{\mathrm{pcd}}_{t'}$ in a web GUI, where they can click a salient point $z_t'$ and specify a waypoint action $w_t'$ by clicking and dragging to rotate or translate a digital twin of the gripper. After the controller $\mathcal{C}$ reaches the waypoint to grasp the train, the process repeats for a waypoint above the bridge. The demonstrator then switches to providing dense actions $a_t$ with a 3D SpaceMouse to carefully place the train on the bridge and tilt it, causing the train to roll.}
\label{fig:interface}
\end{figure}

\subsection{Data Collection Interface for \algabbr}
\label{sec:data-collect}

Without an existing interface that satisfies our need, we design a data collection system to support waypoint specification, salient point annotation, and mode switching seamlessly.
Our hardware setup includes two third-person cameras to provide RGB-D observations to construct a colorized point cloud $o_t^\texttt{pcd}$, and one wrist-mounted camera to provide RGB wrist images $o_t^\texttt{wrist}$. We develop a custom web-based GUI for specifying waypoints and salient points in waypoint mode. To provide dense actions instead, a demonstrator can seamlessly switch from the computer mouse to any dense teleoperation device like a VR/game console controller or a 3D mouse (Spacemouse) as in this work.

The top row of \cref{fig:interface} visualizes the web-based GUI and the process of recording a waypoint action. 
The GUI streams the point cloud of $N$ points $o_t^\texttt{pcd} = \{c_1, c_2, \dots, c_N\}$ to the browser in \textit{real time} and allows a demonstrator to select a salient point $z_t \in \{1, \dots, N\}$ for each waypoint by clicking within the point cloud, (e.g. the small red dot on the toy car next to the mouse cursor.)
After clicking on the salient point, a digital twin of the gripper appears near the salient point to facilitate waypoint specification. 
The demonstrator can set waypoints relative to these salient points with simple click and drag interactions on the virtual gripper.
The salient point specifies the region of interest for interaction while the waypoint captures how to interact with it by specifying 7 DoF target end-effector pose. 
After specifying a waypoint, the linear controller $\mathcal{C}$ defined above interpolates and executes actions to reach the waypoint. Critically, this removes the need for the demonstrator to manually teleoperate long-range movements. The entire waypoint motion is recorded as a sequence $\{(o_t, a_t, \texttt{waypt}, w_{t'}, z_{t'})\}_{t\in \{t', \dots, t'+k_{t'}\}}$ where $t'$ is the timestep when the waypoint is specified and $k_{t'}$ is the number of steps that the controller takes to complete the waypoint $w_{t'}$.

Once the controller finishes executing a waypoint, the demonstrator may specify another waypoint or switch to dense mode for precise manipulation. To take over with dense mode, the demonstrator simply operates the teleoperation device, such as pressing or twisting the joysticks on a controller, and its movements are automatically detected and mapped to delta movements on the end-effector of the robot. This is illustrated by the bottom row of~\cref{fig:interface}, where the operator uses the teleoperation device (Spacemouse in this case) to precisely align and place the toy car onto the narrow bridge. Each step in dense mode is recorded as $(o_t, a_t, \texttt{dense})$. Note that regardless of the mode, we record the full set of observations $o_t$ which includes all camera views as well as proprioception to facilitate data augmentation and to make datasets compatible with any IL policy.

% Notably, all of the above is specified on-the-fly or automatically obtained, without any extra overhead or postprocessing. In fact, comparing to collecting the entire trajectory using teleoperation device only, our system may reduce burden for data collection as it requires the demonstrator to 

\subsection{The Waypoint Policy of \algabbr}
\label{sec:method-waypoint}

The waypoint policy $\pi^\texttt{waypt}$ in \algabbr~takes a point cloud $o_t^\texttt{pcd}$ as input and outputs a 7-DoF end-effector pose $w_t$ for the robot to reach via a controller.
We utilize point clouds as input to cast part of the action prediction problem to learning a salient map over the points. This encourages the policy to attend to the important spatial features (i.e. the handle on a mug) rather than memorize exact locations (i.e. the $x,y,z$ target grasp location). 

\begin{figure}[t]
    \centering
    \includegraphics[width=1.0\textwidth]{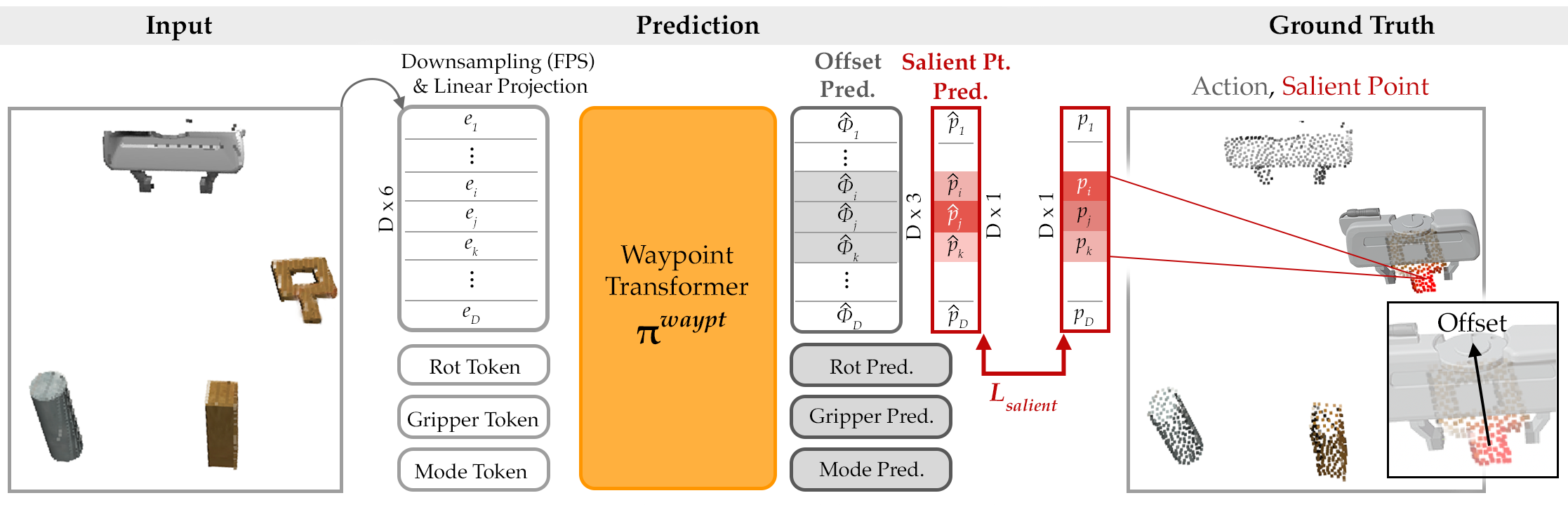}
    \caption{\small \textbf{\algabbr-Waypoint Architecture \& Training Objectives:} \algabbr~takes downsampled point clouds as input, generating per-point tokens $e_i$, and uses a Transformer-style architecture to predict \emph{salient points} and waypoint actions (position, orientation, gripper state). Specifically, \algabbr~predicts the waypoint's positional component as an offset from a salient point. The model outputs a per-point translational offset $\phi_i$, but we only penalize the offset loss on \emph{salient points} (shaded) during training. Salient point prediction is supervised using cross-entropy loss ($L_{\mathrm{salient}}$) between predicted $\hat{p}_i$ and ground truth $p_i$ salient probabilities.}
    \label{fig:method-waypoint}
\end{figure}

The detailed design of the waypoint architecture is illustrated in~\cref{fig:method-waypoint}. At a high level, we would like to have per-point predictions, such as the probability of a point to be salient and the translational offset between the point and the target location of the end-effector, as well as other predictions whose targets are not expressed relative to the points, such as rotation, gripper state, and mode. We use a transformer to process the points and add additional tokens for point-agnostic predictions. 
We first use farthest-point-sampling (FPS)~\citep{pointnet++} to downsamples a raw point cloud to $D=1024$ points, and then convert the points $c_i \in \mathbb{R}^6$ to tokens $e_i \in \mathbb{R}^{d}$ via a shared linear projection layer. 
Then we feed the entire set of tokens into a transformer~\citep{transformer, Radford2019LanguageMA} to get output embeddings. 
Since the points in a point cloud are unordered, the transformer has no positional embedding and does not use a causal mask.
We pass each point embedding through a shared linear layer to get two predictions per point: one for the probability of the point being a salient point $\hat{p}$ and the other for the offset $\hat{\bm{\phi}}_i = (x_i,y_i,z_i)$ between the point position and target waypoint position, illustrated by the middle ``Prediction" panel of \cref{fig:method-waypoint}.

Instead of using a hard one-hot target for salient point prediction, we construct a soft salient map over points where the probability of each point is given by:
\begin{align}
p_i \propto \| c_i - c_k \|_2 \text{ if }  \| c_i - c_k \|_2 \leq r \text{ else } 0
\end{align}
Here, $k$ is the index of the point selected by the user and $r$ is a hyperparameter defining a neighborhood of points that are salient. Within this radius, the probability of saliency decreases with distance to the ground-truth point. In practice, we set to $r$ to 5cm in all of our experiments. 
% Intuitively, this encourages the policy to learn a neighborhood of salient points, where the probability of saliency decreases with distance to the ground-truth point. 
This target distribution of salient points is illustrated by the red points on the right-most panel in~\cref{fig:method-waypoint}, where different shades of red represent the magnitudes of the probability $p_i$.
We train saliency prediction using cross entropy loss on the predicted salient probability $\hat{p}_i$ :
\begin{align}
    L_\text{salient} = -\sum\nolimits_{i} p_i \, \log \hat{p}_i \label{eq:salient}
\end{align}
The goal for the offset prediction is to recover the target end-effector location relative to a predicted salient point, which intuitively aims to ground the waypoint prediction in task-relevant features. This can potentially provide more spatial grounding and awareness that can help with the policy more effectively generalizing. Since we only care about the predicted offset from points with \emph{high} saliency at inference time, we only penalize the offset loss on the points that matter, whose ground-truth salient probabilities $p_i$ are nonzero. Assuming $\bm{\xi}$ is the position of the target end-effector location, $\bm{\varphi}_i$ is the location of the $i$-th point, the loss for offset prediction $\hat{\bm{\phi}}_i$ is:
\begin{align}
    L_\text{offset} = \frac{1}{\sum\nolimits_{i} \mathbbm{1}{\{ p_i > 0 \}}} \sum\nolimits_{i}  \mathbbm{1}{\{ p_i > 0 \}} \cdot \|\bm{\xi} - \bm{\varphi}_i - \hat{\bm{\phi}}_i\|^2 \label{eq:offset}
\end{align}
The prediction targets for rotation, gripper state and next mode are the same regardless of which point is selected. Therefore, we predict them from the output embedding of their respective tokens. 
We use the mean squared error (MSE) loss for rotation on euler angles $L_\text{rot}$, binary cross entropy (BCE) loss for the binary gripper state $L_\text{gripper}$, and negative log-likelihood (NLL) loss for next mode classification $L_\text{mode}$. Since these losses are standard, we move them to~\cref{sec:waypoint-loss} for conciseness. Finally, the full waypoint loss is the sum of all terms, and we find that a simple unweighted sum works well in practice, eliminating the need of additional hyperparameters:
\begin{align}
    L_\text{waypoint} = L_\text{salient} + L_\text{offset} + L_\text{rot} + L_\text{gripper} + L_\text{mode}
\end{align}

By default, the waypoint dataset only consists of $\{(o_t, a_t, \texttt{waypt}, w_t, z_t)\}$ for timesteps $t$ where the demonstrator explicitly specified a waypoint. However, because we use a controller to move the end-effector between its current pose and the waypoint via $k_t$ interpolated actions, we can treat interpolated steps as additional data points to train on with the \emph{same} target waypoint action label. We can expand each observation-action pair to a sequence:
\begin{align}
\{(o_t, a_t, \texttt{waypt}, w_{t'}, z_{t'}) \mid  t' \leq t  \leq t' + \alpha k_{t'}\}
\end{align}
where $\alpha$ is a hyperparameter specifying how much of the interpolated data to train on.

\subsection{The Dense Policy of \algabbr}
% \subsection{\algabbr: The Full Hybrid Policy}
\label{sec:method-dense}

The waypoint policy $\pi^{\texttt{waypt}}(w_t|o_t)$ guides the robot to reach objects in a desired pose. However, long horizon tasks often contain sub-tasks like insertion or alignment that require finer-grained actions. These parts of the task would be easier to accomplish through direct, per-time-step teleoperation, rather than using a series of short waypoints. 
To address this, we train a dense policy, $\pi^{\texttt{dense}}(a_t|o_t^\texttt{wrist})$, which takes over near a salient point to perform precise manipulation before handing control back to the waypoint policy. Note that this policy uses the wrist camera instead of the point cloud $o_t^\texttt{pcd}$, as the dense policy requires high-resolution views that capture close-up object details. Ignoring global observations also helps it to better generalize to different scene arrangements since it only needs to operate locally. 

We instantiate the dense policy of \algabbr~with diffusion policy~\citep{chi2023diffusion} which has been shown to work well in a wide range of manipulation tasks. 
To allow the dense policy to switch back to waypoint mode, we augment its action with an additional mode prediction dimension that predicts the next mode $m_t \in \{0,1,2\}$ corresponding to \{\texttt{waypt}, \texttt{dense}, or \texttt{terminate}\} modes. ~\cref{fig:method-high-level} illustrates how the dense policy fits into the entire \algabbr~framework.
We train the dense policy using the entire dataset and use the interpolated steps to augment our data. 
This provides the policy with more data and encourages it to be robust to slightly early or late mode switches.

\vspace{-0.2cm}
\section{Experiments}
\label{sec:experiments}

In this section, we evaluate how \algabbr's attention to salient points and hybrid policy architecture impacts its performance and generalization on a suite of four challenging real-world tasks and two simulated manipulation tasks. In all experiments, we assume access to two external camera viewpoints and a wrist-mounted camera on the Franka Panda robot arm. See~\cref{sec:impl-waypoint} for the implementation details.

\subsection{Evaluation on Precise and Long-Horizon Tasks} 
We first evaluate \algabbr's performance on complex, long-horizon tasks that demand precision. We hypothesize that by interleaving waypoint actions predicted from point clouds, and dense actions predicted from close-up wrist-camera images once near a salient point, \algabbr~will more effectively be able to complete this class of tasks compared to baselines which do not exploit salient points or a hybrid mode-switching policy. To assess this, we consider 3 challenging real-world tasks which we found to be infeasible to teleoperate in waypoint-mode alone due to the required degree of precision and reactivity: \emph{Cup Stack} (30 demonstrations), \emph{Train Track} (30 demonstrations), and \emph{Coffee Making} (60 demonstrations). See \cref{fig:rollouts} for visualizations of each task and example rollouts. For each task, we consider a large and diverse space of initial configurations (\cref{fig:state_distribution}), varying the relative locations of objects (cups, mugs, coffee pods, machine, train track).

\paragraph{Baselines:} We compare against three baselines, the first being \textbf{Diffusion Policy (DP)} from~\cite{chi2023diffusion} with images from all three cameras as input. We intend for this baseline to demonstrate the benefit of waypoint modes (or lack thereof) for challenging manipulation tasks. The second baseline is \textbf{HYDRA} ~\citep{belkhale2023hydra}, a hybrid IL policy which takes images from all three views as input and outputs waypoint and dense actions. Given that it is image-based, HYDRA uses a \emph{multiheaded} policy with a shared image encoder as input to waypoint, dense, and mode prediction heads. Its waypoint prediction head outputs a waypoint \emph{without} any intermediate salient representation. We choose this baseline to demonstrate the effects of separate input modalities (point cloud vs. wrist image) for waypoint and dense modes, as well as the benefit of using salient points to ground waypoint actions. The original HYDRA implementation used a simple MLP for the dense head, but we update it to a diffusion policy for fair comparison. The last baseline is \textbf{Fine-tuned OpenVLA}, where we fine-tune the recent OpenVLA model~\citep{kim2024openvla}, pretrained on large robotics datasets, on our single-task datasets. We use it as an independent baseline given the recent trend of leveraging large prior datasets towards generalizable robotic manipulation. Due to its design restrictions, this model can only take a single third-person image as input. 

\begin{figure}[t]
\centering
\includegraphics[width=1.0\textwidth]{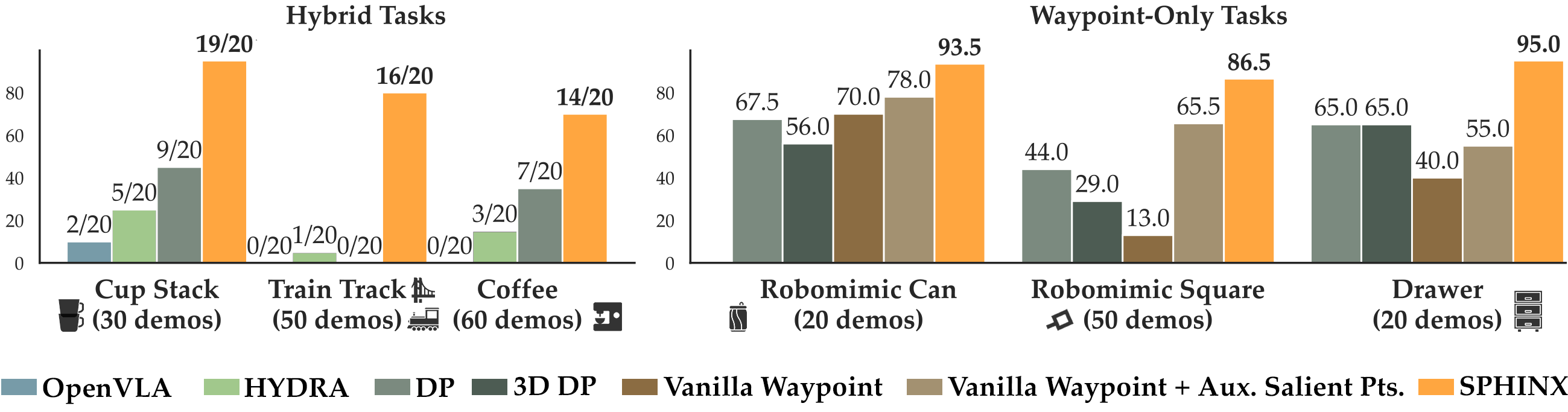}
\caption{\small \textbf{Success Rates Across Tasks:} Left: \algabbr~outperforms image-only dense baselines (OpenVLA, diffusion policy) as well as a hybrid baseline (HYDRA) across 3 challenging real-world tasks (\cref{fig:rollouts}) collected with \emph{hybrid} mode teleoperation. \emph{Train Track} requires a degree of precision that baselines lack, while \algabbr's use of salient points and hybrid actions enables precise, long-horizon manipulation. Right: \algabbr~performs $1.6\times$ better than the SoTA image or point-cloud based diffusion policies across tasks teleoperated in only \emph{waypoint} mode. Comparisons with the two vanilla waypoint baselines also show that both saliency prediction and the relative waypoint action representation contribute to \algabbr's strong performance.
}
\label{fig:hybrid_waypoint_results}
\end{figure}

In \cref{fig:hybrid_waypoint_results} (left), \algabbr~achieves the best performance across all three challenging tasks. Fine-tuned OpenVLA struggles to achieve the required precision, lacking mode switches and close-up wrist images. HYDRA shows nonzero performance but suffers from inaccurate waypoint predictions without salient point attention and point clouds. Diffusion policy performs the best among the baselines but struggles to generalize across initial configurations without the waypoint mode. 
As shown in~\cref{fig:state_distribution}, \algabbr~generalizes better across various object placements, while baselines tend to memorize a few arrangements. Overall, these tasks are highly unforgiving of grasping failures and imprecision. Baseline methods particularly struggle to make progress in the train and coffee tasks where early mistakes (missed grasps or placements with the mug, coffee pod, train) derail an entire rollout. \algabbr's use of salient point attention and close-up wrist images allows the policy to carefully proceed through difficult task phases, leading to higher success across scenes.

\begin{figure}[t]
\vspace{-0.3cm}
\centering
\includegraphics[width=1.0\textwidth]{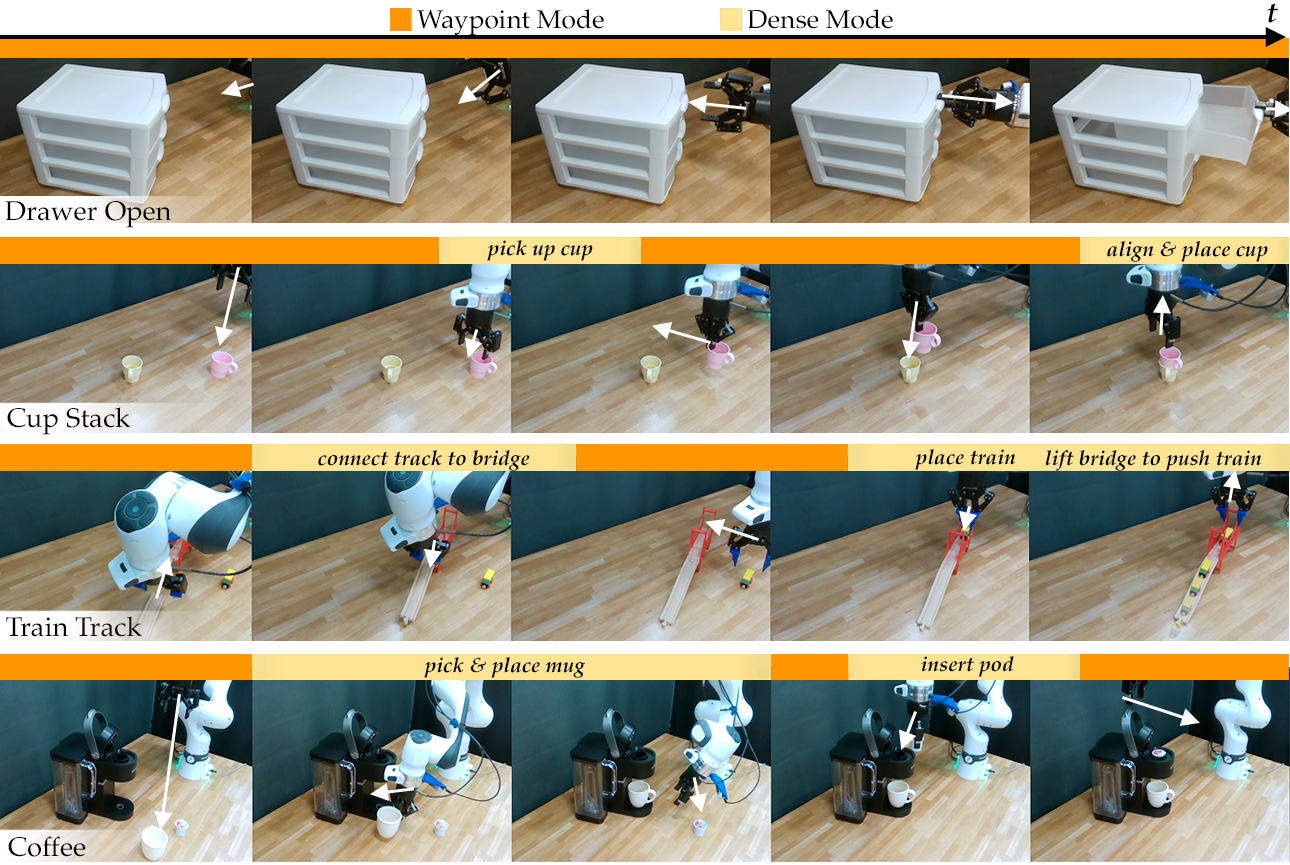}
\caption{\small \textbf{\algabbr~Rollouts}: We evaluate \algabbr~across a suite of challenging real-world tasks subject to wide initial state variations. \algabbr's waypoint mode alone is precise enough to handle tasks like drawer opening, while the full hybrid policy leverages different action modes to tackle complex tasks such as cup stacking, building and playing with a toy train set, and making coffee.}
\label{fig:rollouts}
\end{figure}

\subsection{Waypoint Policy Ablations}
\label{sec:waypt_exps}
Although \algabbr's performance in challenging real-world tasks relies on the dense policy, the waypoint policy is crucial for its strong performance and generalization as it reliably guides the robot to task-relevant locations.  
In this section, we validate the design choices behind \algabbr's waypoint policy through ablations and comparisons against state-of-the-art (SoTA) IL policies. To isolate the impact of the waypoint policy from that of the dense policy, we conduct these experiments exclusively on tasks that can be teleoperated solely in \emph{waypoint mode}, without the use of dense mode.

\begin{figure}[!htbp]
\vspace{-0.3cm}
\centering
\includegraphics[width=1.0\textwidth]{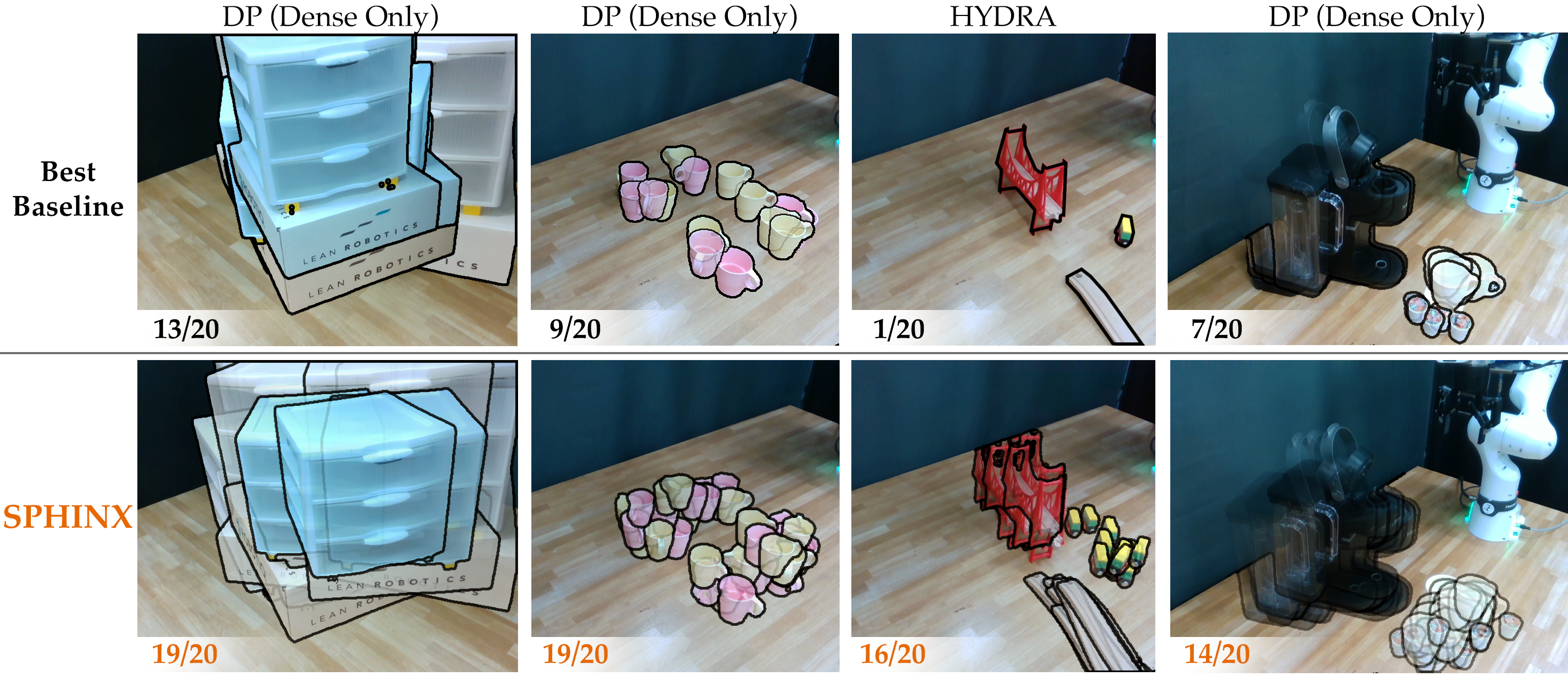}
\caption{\small \textbf{Distribution of Successful Rollouts}: Across 4 real-world tasks, we visualize the initial state distribution for successful trials across methods by overlaying segmented initial images. Notably, \algabbr~handles the widest degree of spatial variety, while achieving a much higher task success rate compared to the most competitive baseline. For particularly difficult tasks like \emph{Train Track} and \emph{Coffee}, the baselines tend to memorize motions for particular object arrangements (i.e. mug behind pods) rather than generalize to spatial variations.}
\label{fig:state_distribution}
\end{figure}

We posit that \algabbr~can achieve a higher task success rate across a wide range of initial spatial configurations by effectively using waypoints (via salient points and offsets) to reduce the action prediction horizon and maintain higher precision and better spatial awareness than alternatives.
We consider three tasks, one real task of opening an articulated drawer (\emph{Drawer}, 20 demonstrations) and two simulated environments in Robomimic~\citep{mandlekar2021what} (\emph{Can}, 20 demonstrations, and \emph{Square}, 50 demonstrations). 

\textbf{Baselines:}
We compare \algabbr~against four baselines. 
The first two are SoTA IL policies in robotics --- dense-only image-based \textbf{Diffusion Policy}~\citep{chi2023diffusion} and its point-cloud based extension \textbf{3D Diffusion Policy (3D DP)}~\citep{ze20243d}.  We train them on the interpolated data $\{o_t, a_t\}$. In addition to RGB images from the same external cameras used by the point cloud-based methods, the image-based DP also incorporates wrist camera inputs to ensure optimal performance. 

Additionally, we consider two waypoint baselines that can also be seen as ablations. \textbf{Vanilla Waypoint} uses the same input and Transformer backbone as \algabbr~but it removes the salient point prediction and offset prediction. It instead adds a Translation Token, similar to the Rotation Token, to the Transformer and predicts the target translation from the output of that token using MSE loss. \textbf{Vanilla Waypoint + Auxiliary Salient Points} predicts the translation the same way as vanilla waypoint but adds the salient point prediction of \algabbr~as an auxiliary task. It can also be viewed as \algabbr~without offset prediction. We choose the two waypoint baselines to ablate over the importance of salient point and offset prediction separately.

\textbf{Results:} 
In \cref{fig:hybrid_waypoint_results} (right), we find that  \algabbr~achieves $\mathbf{1.6\times}$ better performance than the best baseline between SoTA image-based or 3D diffusion policies. By carefully inspecting the two vanilla waypoint variants, we can see that vanilla waypoint is only slightly better in the simplest \emph{Can} task but notably worse than the diffusion policies on the more complex simulation task of \emph{Square} as well as the real world \emph{Drawer} task. 
This is interesting as it suggests that the shorter prediction horizon of waypoint policies alone is not necessarily a benefit without predicting salient points. Adding the salient point prediction as an auxiliary task improves the performance of vanilla waypoint without changing the action representation, especially in \emph{Square} which requires highly precise maneuvers like grasping and hanging a small tool handle where attention to specific object parts is crucial. This suggests the utility of salient points for encouraging better action prediction. \algabbr's dominant performance over the two waypoint baselines suggests the effectiveness of anchoring waypoint action prediction \emph{relative} to \emph{salient points}.

% \begin{table}[t]
% \centering
% \begin{tabular}{l c c c c}
% \toprule
% \multirow{2}{*}{Method}             & \multirow{2}{*}{Input} & Can & Square & Drawer \\
% & & (20 demos) & (50 demos) & (20 demos) \\
% \midrule
% Diffusion Policy~\citep{chi2023diffusion} & Image  &   67.5\%  &  44\% & 65\%  \\
% 3D Diffusion Policy~\citep{ze20243d} & Pointcloud  & 56\% & 29\% & 65\%\\
% \midrule
% Vanilla Waypoint  & Pointcloud & 70\% &  13\% & 40\% \\
% Vanilla Waypoint + Aux. Salient Points & Pointcloud &78\% & 65.5\% & 55\% \\
% \cellcolor{orange!30}\algabbr  & \cellcolor{orange!30}Pointcloud & \cellcolor{orange!30}\textbf{93.5\%} &    \cellcolor{orange!30}\textbf{86.5\%} & \cellcolor{orange!30} \textbf{95\%}\\
% \bottomrule
% \end{tabular}
% \caption{\small \textbf{Waypoint-Only Task Evaluations}: \algabbr~performs $1.6\times$ better than the SoTA image or point-cloud based Diffusion Policies. Comparisons with the two vanilla waypoint baselines also show that both saliency prediction and the relative waypoint action representation contribute to \algabbr~strong performance.}
% \label{tab:waypoint_results}
% \end{table}

\begin{table}[t]
\centering
\begin{tabular}{l c c c c}
\toprule
\multirow{2}{*}{Method}             & Drawer & Drawer & Cup Stack & Cup Stack \\
& Unseen height & Novel viewpoint & Visual distraction & Novel viewpoint \\
\midrule
Diffusion Policy & 4/10 &   1/10 &  1/10 & 0/10  \\
\cellcolor{orange!30}\algabbr  & \cellcolor{orange!30} \textbf{9/10} & \cellcolor{orange!30} \textbf{9/10} &  \cellcolor{orange!30}\textbf{8/10} & \cellcolor{orange!30}\textbf{9/10} \\
\bottomrule
\end{tabular}
\caption{\small \textbf{OOD Results}: 
We compare \algabbr~to DP on \emph{Drawer} and \emph{Cup Stack} across various unseen scenarios. In \emph{Drawer}, \algabbr~successfully opens the drawer at unseen heights by attending to the handle as a salient point, while DP struggles. In \emph{Cup Stack}, \algabbr~maintains strong performance despite visual distractors, as its salient points focus on the cups, and the wrist-only dense policy ignores surrounding scene changes. Finally, \algabbr~generalizes to unseen camera viewpoints in both tasks, whereas DP's image-based approach suffers.}
\label{tab:ood-gen}
\end{table}

\subsection{Visual Generalization}

\begin{figure}[t]
\centering
\includegraphics[width=1.0\textwidth]{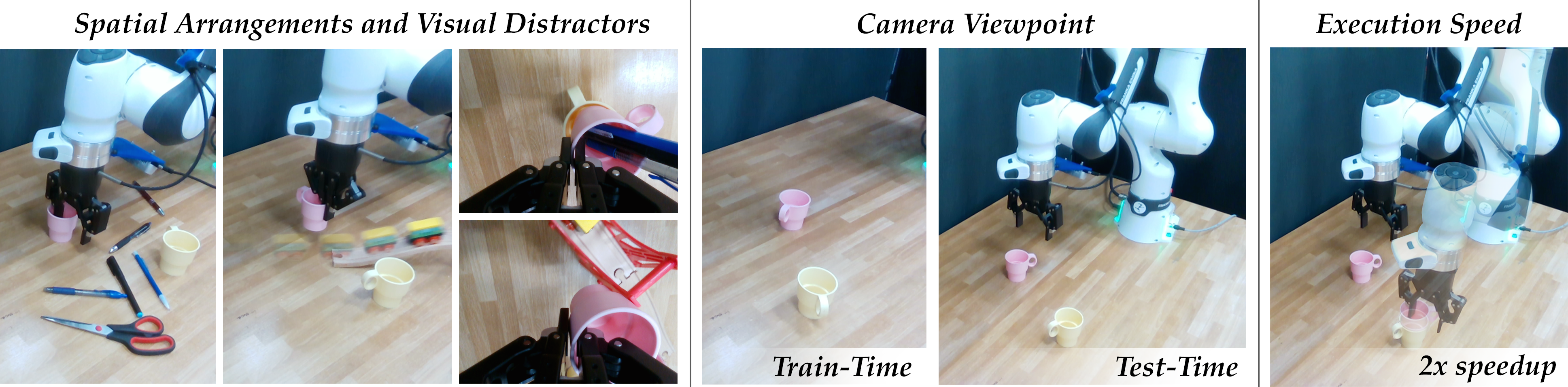}
\caption{\small \textbf{Generalization Capabilities}: \algabbr~generalizes across multiple axes on \emph{Cup Stack}, handling static (e.g., office supplies) and dynamic (e.g., moving toy train) distractors (see \href{http://sphinx-manip.github.io}{website}). Using calibrated point clouds as input further allows for generalization to unseen camera viewpoints at test time, and the waypoint controller enables $accelerated$ execution speeds, completing \emph{Cup Stack} in 9.2 s. on average.}
\label{fig:generalization}
\end{figure}

We next evaluate \algabbr's ability to handle \emph{visual} rather than only spatial generalization on \emph{Drawer} and \emph{Cup Stack}. As seen in \cref{tab:ood-gen}, \algabbr~demonstrates a promising degree of generalization to \underline{\emph{visual distractors}} during execution, and \underline{\emph{unseen third-person camera viewpoints}} (\cref{fig:generalization}),  retaining high performance. This is likely enabled by using viewpoint-agnostic point clouds  (assuming calibration), salient points encouraging the policy to ignore distractors, and wrist-camera images in dense mode which are largely unaffected by changes in the surrounding scene. Diffusion policy heavily overfits to the training scenes and suffers a noticeable performance drop.

\subsection{Execution Speed}
A key advantage of \algabbr~over dense-only methods is its decoupled waypoint and dense policies. While dense methods are tied to the speed of actions recorded during data collection, \algabbr~uses a waypoint controller that allows flexible execution speeds at test time. We specifically collect all data across all tasks using a controller limited to a maximum delta of 1 cm at 10Hz. After training \algabbr~and dense-only diffusion policy on \emph{Cup Stack}, we perform 10 trials of the task, where we compare \algabbr~implemented with a $2\times$ sped-up controller (2 cm maximum delta at test-time) to DP trained on the $1\times$ data. \algabbr~completes the task in an average of \textbf{$\mathbf{9.2}$ seconds}, a $1.7\times$ speedup over diffusion policy ($15.6$ seconds). While further speed increases led to controller imprecision, \algabbr~ has potential for even faster execution on more capable hardware.

\section{Limitations and Conclusion}
\label{sec:conclusion}
% \vspace{-0.3cm}

% \section{Limitations and Failure Modes}
\algabbr~demonstrates strong performance and generalization across a range of tasks, but our policy is not without failures. The majority of \algabbr's failures stem from the dense policy being slightly imprecise for grasping or manipulation. Although we mitigate this by using the dense policy only for short horizons near salient points, performing the ``last mile'' of precise insertion or alignment remains challenging for some tasks. Additionally, our data collection interface uses a linear controller to reach waypoints. This currently limits \algabbr~to fairly quasistatic tasks without fast, dynamic movements. Finally, the performance of our waypoint policy is limited by the quality of the input point cloud. We currently perform a one-time calibration procedure to obtain multi-view extrinsics and point clouds, but sensor noise and calibration error is not completely avoidable.

To summarize, we present \algabbr, a visuomotor IL policy which learns to perform complex manipulation tasks from a limited amount of demonstrations while generalizing across many axes: novel spatial arrangements, visual distractors, novel viewpoints, and even customizable execution speeds (\cref{fig:generalization}). \algabbr~achieves this using a hybrid policy architecture that takes point clouds and wrist-images as input, and outputs waypoints and dense actions guided by salient points. \algabbr~achieves an average success rate of $\mathbf{86.6\%}$ across 2 simulated and 4 real-world high-precision, long-horizon tasks, including making coffee and assembling a train set with several pieces. Our policy outperforms state-of-the-art IL baselines by $\mathbf{41.1\%}$ on average across $\mathbf{440}$ real world robot trials. Avenues for future work include improving dense mode manipulation with additional sensing modalities such as tactile sensors, extending our data collection interface to support dynamic manipulation tasks, and deploying \algabbr~in the wild across a range of real-world environments.

% \newpage
% \input{tex/5-v2.tex}

% funding sources go here

% The acknowledgments are automatically included only in the final version of the paper.
% \footnotesize
\subsection*{Acknowledgments}
This work is in part supported by funds from NSF Awards 2132847 and 2218760, the Office of Naval Research under ONR N00014-22-1-2293, DARPA TIAMAT, DARPA \#W911NF2210214, as well as Intrinsic and the Stanford Institute for Human-Centered Artificial Intelligence (HAI). Priya Sundaresan is supported by an NSF GRFP. We would like to thank Suvir Mirchandani, Suneel Belkhale and Yuchen Cui for their helpful feedback and suggestions.
% \normalsize

\newpage
\bibliography{iclr2024_conference}

\begin{thebibliography}{37}
\providecommand{\natexlab}[1]{#1}
\providecommand{\url}[1]{\texttt{#1}}
\expandafter\ifx\csname urlstyle\endcsname\relax
  \providecommand{\doi}[1]{doi: #1}\else
  \providecommand{\doi}{doi: \begingroup \urlstyle{rm}\Url}\fi

\bibitem[Agia et~al.(2022)Agia, Migimatsu, Wu, and Bohg]{agia2022taps}
Christopher Agia, Toki Migimatsu, Jiajun Wu, and Jeannette Bohg.
\newblock Taps: Task-agnostic policy sequencing.
\newblock \emph{arXiv preprint arXiv:2210.12250}, 2022.

\bibitem[Ahn et~al.(2022)Ahn, Brohan, Brown, Chebotar, Cortes, David, Finn, Fu, Gopalakrishnan, Hausman, Herzog, Ho, Hsu, Ibarz, Ichter, Irpan, Jang, Ruano, Jeffrey, Jesmonth, Joshi, Julian, Kalashnikov, Kuang, Lee, Levine, Lu, Luu, Parada, Pastor, Quiambao, Rao, Rettinghouse, Reyes, Sermanet, Sievers, Tan, Toshev, Vanhoucke, Xia, Xiao, Xu, Xu, Yan, and Zeng]{saycan2022arxiv}
Michael Ahn, Anthony Brohan, Noah Brown, Yevgen Chebotar, Omar Cortes, Byron David, Chelsea Finn, Chuyuan Fu, Keerthana Gopalakrishnan, Karol Hausman, Alex Herzog, Daniel Ho, Jasmine Hsu, Julian Ibarz, Brian Ichter, Alex Irpan, Eric Jang, Rosario~Jauregui Ruano, Kyle Jeffrey, Sally Jesmonth, Nikhil Joshi, Ryan Julian, Dmitry Kalashnikov, Yuheng Kuang, Kuang-Huei Lee, Sergey Levine, Yao Lu, Linda Luu, Carolina Parada, Peter Pastor, Jornell Quiambao, Kanishka Rao, Jarek Rettinghouse, Diego Reyes, Pierre Sermanet, Nicolas Sievers, Clayton Tan, Alexander Toshev, Vincent Vanhoucke, Fei Xia, Ted Xiao, Peng Xu, Sichun Xu, Mengyuan Yan, and Andy Zeng.
\newblock Do as i can and not as i say: Grounding language in robotic affordances.
\newblock In \emph{arXiv preprint arXiv:2204.01691}, 2022.

\bibitem[Atkeson \& Schaal(1997)Atkeson and Schaal]{atkeson1997robot}
Christopher~G Atkeson and Stefan Schaal.
\newblock Robot learning from demonstration.
\newblock In \emph{ICML}, volume~97, pp.\  12--20, 1997.

\bibitem[Belkhale et~al.(2023)Belkhale, Cui, and Sadigh]{belkhale2023hydra}
Suneel Belkhale, Yuchen Cui, and Dorsa Sadigh.
\newblock Hydra: Hybrid robot actions for imitation learning.
\newblock In \emph{Conference on Robot Learning}, pp.\  2113--2133. PMLR, 2023.

\bibitem[Bharadhwaj et~al.(2024)Bharadhwaj, Vakil, Sharma, Gupta, Tulsiani, and Kumar]{bharadhwaj2024roboagent}
Homanga Bharadhwaj, Jay Vakil, Mohit Sharma, Abhinav Gupta, Shubham Tulsiani, and Vikash Kumar.
\newblock Roboagent: Generalization and efficiency in robot manipulation via semantic augmentations and action chunking.
\newblock In \emph{2024 IEEE International Conference on Robotics and Automation (ICRA)}, pp.\  4788--4795. IEEE, 2024.

\bibitem[Black et~al.()Black, Brown, Driess, Esmail, Equi, Finn, Fusai, Groom, Hausman, Ichter, Jakubczak, Jones, Ke, Levine, Li-Bell, Mothukuri, Nair, Pertsch, Shi, and Tanner]{pi0}
Kevin Black, Noah Brown, Danny Driess, Adnan Esmail, Michael Equi, Chelsea Finn, Niccolo Fusai, Lachy Groom, Karol Hausman, Brian Ichter, Szymon Jakubczak, Tim Jones, Liyiming Ke, Sergey Levine, Adrian Li-Bell, Mohith Mothukuri, Suraj Nair, Karl Pertsch, Lucy Shi, and James Tanner.
\newblock $\pi$0 : A vision-language-action flow model for general robot control physical intelligence.
\newblock URL \url{https://www.physicalintelligence.company/download/pi0.pdf}.

\bibitem[Borja-Diaz et~al.(2022)Borja-Diaz, Mees, Kalweit, Hermann, Boedecker, and Burgard]{Affordance-Learning-from-Play}
Jessica Borja-Diaz, Oier Mees, Gabriel Kalweit, Lukas Hermann, Joschka Boedecker, and Wolfram Burgard.
\newblock Affordance learning from play for sample-efficient policy learning.
\newblock In \emph{2022 International Conference on Robotics and Automation (ICRA)}, pp.\  6372--6378, 2022.
\newblock \doi{10.1109/ICRA46639.2022.9811889}.

\bibitem[Chi et~al.(2023)Chi, Feng, Du, Xu, Cousineau, Burchfiel, and Song]{chi2023diffusion}
Cheng Chi, Siyuan Feng, Yilun Du, Zhenjia Xu, Eric Cousineau, Benjamin Burchfiel, and Shuran Song.
\newblock Diffusion policy: Visuomotor policy learning via action diffusion.
\newblock \emph{arXiv preprint arXiv:2303.04137}, 2023.

\bibitem[Chi et~al.(2024)Chi, Xu, Pan, Cousineau, Burchfiel, Feng, Tedrake, and Song]{chi2024universal}
Cheng Chi, Zhenjia Xu, Chuer Pan, Eric Cousineau, Benjamin Burchfiel, Siyuan Feng, Russ Tedrake, and Shuran Song.
\newblock Universal manipulation interface: In-the-wild robot teaching without in-the-wild robots.
\newblock \emph{arXiv preprint arXiv:2402.10329}, 2024.

\bibitem[Dalal et~al.(2021)Dalal, Pathak, and Salakhutdinov]{dalal2021accelerating}
Murtaza Dalal, Deepak Pathak, and Russ~R Salakhutdinov.
\newblock Accelerating robotic reinforcement learning via parameterized action primitives.
\newblock \emph{Advances in Neural Information Processing Systems}, 34:\penalty0 21847--21859, 2021.

\bibitem[Gervet et~al.(2023)Gervet, Xian, Gkanatsios, and Fragkiadaki]{gervet2023act3d3dfeaturefield}
Theophile Gervet, Zhou Xian, Nikolaos Gkanatsios, and Katerina Fragkiadaki.
\newblock Act3d: 3d feature field transformers for multi-task robotic manipulation, 2023.
\newblock URL \url{https://arxiv.org/abs/2306.17817}.

\bibitem[Goyal et~al.(2023)Goyal, Xu, Guo, Blukis, Chao, and Fox]{goyal2023rvt}
Ankit Goyal, Jie Xu, Yijie Guo, Valts Blukis, Yu-Wei Chao, and Dieter Fox.
\newblock Rvt: Robotic view transformer for 3d object manipulation.
\newblock In \emph{Conference on Robot Learning}, pp.\  694--710. PMLR, 2023.

\bibitem[He et~al.(2016)He, Zhang, Ren, and Sun]{he2016deep}
Kaiming He, Xiangyu Zhang, Shaoqing Ren, and Jian Sun.
\newblock Deep residual learning for image recognition.
\newblock In \emph{Proceedings of the IEEE conference on computer vision and pattern recognition}, pp.\  770--778, 2016.

\bibitem[Jedrzej~Orbik(2021)]{OrbikEbert2021OculusReader}
Frederik~Ebert Jedrzej~Orbik.
\newblock Oculus reader: Robotic teleoperation interface, 2021.
\newblock URL \url{https://github.com/rail-berkeley/oculus_reader}.
\newblock Accessed: 2024-09-20.

\bibitem[Kim et~al.(2024)Kim, Pertsch, Karamcheti, Xiao, Balakrishna, Nair, Rafailov, Foster, Lam, Sanketi, et~al.]{kim2024openvla}
Moo~Jin Kim, Karl Pertsch, Siddharth Karamcheti, Ted Xiao, Ashwin Balakrishna, Suraj Nair, Rafael Rafailov, Ethan Foster, Grace Lam, Pannag Sanketi, et~al.
\newblock Openvla: An open-source vision-language-action model.
\newblock \emph{arXiv preprint arXiv:2406.09246}, 2024.

\bibitem[Kingma \& Ba(2015)Kingma and Ba]{kingma2014adam}
Diederik~P. Kingma and Jimmy Ba.
\newblock Adam: {A} method for stochastic optimization.
\newblock In \emph{International Conference on Learning Representations (ICLR)}, 2015.

\bibitem[Mandlekar et~al.(2021)Mandlekar, Xu, Wong, Nasiriany, Wang, Kulkarni, Fei{-}Fei, Savarese, Zhu, and Mart{\'{\i}}n{-}Mart{\'{\i}}n]{mandlekar2021what}
Ajay Mandlekar, Danfei Xu, Josiah Wong, Soroush Nasiriany, Chen Wang, Rohun Kulkarni, Li~Fei{-}Fei, Silvio Savarese, Yuke Zhu, and Roberto Mart{\'{\i}}n{-}Mart{\'{\i}}n.
\newblock What matters in learning from offline human demonstrations for robot manipulation.
\newblock In \emph{Conference on Robot Learning (CoRL)}, 2021.

\bibitem[Nasiriany et~al.(2022)Nasiriany, Liu, and Zhu]{nasiriany2022augmenting}
Soroush Nasiriany, Huihan Liu, and Yuke Zhu.
\newblock Augmenting reinforcement learning with behavior primitives for diverse manipulation tasks.
\newblock In \emph{2022 International Conference on Robotics and Automation (ICRA)}, pp.\  7477--7484. IEEE, 2022.

\bibitem[Pomerleau(1988)]{pomerleau1988alvinn}
Dean~A Pomerleau.
\newblock Alvinn: An autonomous land vehicle in a neural network.
\newblock \emph{Advances in neural information processing systems}, 1, 1988.

\bibitem[Qi et~al.(2017)Qi, Yi, Su, and Guibas]{pointnet++}
Charles~R. Qi, Li~Yi, Hao Su, and Leonidas~J. Guibas.
\newblock Pointnet++: deep hierarchical feature learning on point sets in a metric space.
\newblock In \emph{Proceedings of the 31st International Conference on Neural Information Processing Systems}, NIPS'17, pp.\  5105–5114, Red Hook, NY, USA, 2017. Curran Associates Inc.
\newblock ISBN 9781510860964.

\bibitem[Radford et~al.(2019)Radford, Wu, Child, Luan, Amodei, and Sutskever]{Radford2019LanguageMA}
Alec Radford, Jeff Wu, Rewon Child, David Luan, Dario Amodei, and Ilya Sutskever.
\newblock Language models are unsupervised multitask learners.
\newblock 2019.
\newblock URL \url{https://api.semanticscholar.org/CorpusID:160025533}.

\bibitem[Reuss et~al.(2023)Reuss, Li, Jia, and Lioutikov]{reuss2023goal-diffuse}
Moritz Reuss, Maximilian Li, Xiaogang Jia, and Rudolf Lioutikov.
\newblock Goal conditioned imitation learning using score-based diffusion policies.
\newblock In \emph{Robotics: Science and Systems}, 2023.

\bibitem[Saxena et~al.(2008)Saxena, Driemeyer, and Ng]{grasp-point}
Ashutosh Saxena, Justin Driemeyer, and Andrew~Y. Ng.
\newblock Robotic grasping of novel objects using vision.
\newblock \emph{The International Journal of Robotics Research}, 27\penalty0 (2):\penalty0 157--173, 2008.
\newblock \doi{10.1177/0278364907087172}.
\newblock URL \url{https://doi.org/10.1177/0278364907087172}.

\bibitem[Schaal(1996)]{schaal1996learning}
Stefan Schaal.
\newblock Learning from demonstration.
\newblock \emph{Advances in neural information processing systems}, 9, 1996.

\bibitem[Shi et~al.(2023)Shi, Sharma, Zhao, and Finn]{shi2023waypoint}
Lucy~Xiaoyang Shi, Archit Sharma, Tony~Z Zhao, and Chelsea Finn.
\newblock Waypoint-based imitation learning for robotic manipulation.
\newblock \emph{arXiv preprint arXiv:2307.14326}, 2023.

\bibitem[Shridhar et~al.(2022)Shridhar, Manuelli, and Fox]{shridhar2022peract}
Mohit Shridhar, Lucas Manuelli, and Dieter Fox.
\newblock Perceiver-actor: A multi-task transformer for robotic manipulation.
\newblock In \emph{Proceedings of the 6th Conference on Robot Learning (CoRL)}, 2022.

\bibitem[Shridhar et~al.(2023)Shridhar, Manuelli, and Fox]{shridhar2023perceiver}
Mohit Shridhar, Lucas Manuelli, and Dieter Fox.
\newblock Perceiver-actor: A multi-task transformer for robotic manipulation.
\newblock In \emph{Conference on Robot Learning}, pp.\  785--799. PMLR, 2023.

\bibitem[Sundaresan et~al.(2023)Sundaresan, Belkhale, Sadigh, and Bohg]{sundaresan2023kite}
Priya Sundaresan, Suneel Belkhale, Dorsa Sadigh, and Jeannette Bohg.
\newblock Kite: Keypoint-conditioned policies for semantic manipulation.
\newblock \emph{arXiv preprint arXiv:2306.16605}, 2023.

\bibitem[Sundermeyer et~al.(2021)Sundermeyer, Mousavian, Triebel, and Fox]{sundermeyer2021contact}
Martin Sundermeyer, Arsalan Mousavian, Rudolph Triebel, and Dieter Fox.
\newblock Contact-graspnet: Efficient 6-dof grasp generation in cluttered scenes.
\newblock In \emph{2021 IEEE International Conference on Robotics and Automation (ICRA)}, pp.\  13438--13444. IEEE, 2021.

\bibitem[Vaswani et~al.(2017)Vaswani, Shazeer, Parmar, Uszkoreit, Jones, Gomez, Kaiser, and Polosukhin]{transformer}
Ashish Vaswani, Noam Shazeer, Niki Parmar, Jakob Uszkoreit, Llion Jones, Aidan~N Gomez, \L~ukasz Kaiser, and Illia Polosukhin.
\newblock Attention is all you need.
\newblock In I.~Guyon, U.~Von Luxburg, S.~Bengio, H.~Wallach, R.~Fergus, S.~Vishwanathan, and R.~Garnett (eds.), \emph{Advances in Neural Information Processing Systems}, volume~30. Curran Associates, Inc., 2017.
\newblock URL \url{https://proceedings.neurips.cc/paper_files/paper/2017/file/3f5ee243547dee91fbd053c1c4a845aa-Paper.pdf}.

\bibitem[Wang et~al.(2024)Wang, Hart, Surovik, Kelestemur, Huang, Zhao, Yeatman, Wang, Walters, and Platt]{wang2024equivariant}
Dian Wang, Stephen Hart, David Surovik, Tarik Kelestemur, Haojie Huang, Haibo Zhao, Mark Yeatman, Jiuguang Wang, Robin Walters, and Robert Platt.
\newblock Equivariant diffusion policy.
\newblock \emph{arXiv preprint arXiv:2407.01812}, 2024.

\bibitem[Yang et~al.(2024{\natexlab{a}})Yang, Cao, Deng, Antonova, Song, and Bohg]{yang2024equibot}
Jingyun Yang, Zi-ang Cao, Congyue Deng, Rika Antonova, Shuran Song, and Jeannette Bohg.
\newblock Equibot: Sim (3)-equivariant diffusion policy for generalizable and data efficient learning.
\newblock \emph{arXiv preprint arXiv:2407.01479}, 2024{\natexlab{a}}.

\bibitem[Yang et~al.(2024{\natexlab{b}})Yang, Deng, Wu, Antonova, Guibas, and Bohg]{yang2024equivact}
Jingyun Yang, Congyue Deng, Jimmy Wu, Rika Antonova, Leonidas Guibas, and Jeannette Bohg.
\newblock Equivact: Sim (3)-equivariant visuomotor policies beyond rigid object manipulation.
\newblock In \emph{2024 IEEE International Conference on Robotics and Automation (ICRA)}, pp.\  9249--9255. IEEE, 2024{\natexlab{b}}.

\bibitem[Yu et~al.(2023)Yu, Xiao, Stone, Tompson, Brohan, Wang, Singh, Tan, Peralta, Ichter, et~al.]{yu2023scaling}
Tianhe Yu, Ted Xiao, Austin Stone, Jonathan Tompson, Anthony Brohan, Su~Wang, Jaspiar Singh, Clayton Tan, Jodilyn Peralta, Brian Ichter, et~al.
\newblock Scaling robot learning with semantically imagined experience.
\newblock \emph{arXiv preprint arXiv:2302.11550}, 2023.

\bibitem[Ze et~al.(2024)Ze, Zhang, Zhang, Hu, Wang, and Xu]{ze20243d}
Yanjie Ze, Gu~Zhang, Kangning Zhang, Chenyuan Hu, Muhan Wang, and Huazhe Xu.
\newblock 3d diffusion policy.
\newblock \emph{arXiv preprint arXiv:2403.03954}, 2024.

\bibitem[Zhang et~al.(2024)Zhang, Hu, You, and Gao]{zhang2024leveraging}
Tong Zhang, Yingdong Hu, Jiacheng You, and Yang Gao.
\newblock Leveraging locality to boost sample efficiency in robotic manipulation.
\newblock \emph{arXiv preprint arXiv:2406.10615}, 2024.

\bibitem[Zhao et~al.(2023)Zhao, Kumar, Levine, and Finn]{zhao2023learning}
Tony~Z Zhao, Vikash Kumar, Sergey Levine, and Chelsea Finn.
\newblock Learning fine-grained bimanual manipulation with low-cost hardware.
\newblock \emph{arXiv preprint arXiv:2304.13705}, 2023.

\end{thebibliography}
\bibliographystyle{iclr2024_conference}

\newpage
\appendix
\section{Additional Losses for the Waypoint Policy in \algabbr}
\label{sec:waypoint-loss}
In~\cref{sec:method-waypoint} we define the salient point prediction loss and offset loss for \algabbr. 
Here we complete the full loss definitions for the waypoint policy. 
Recall that the waypoint policy needs to predict translation $\hat{\bm{\xi}}$, rotation $\hat{\bm{\alpha}}$, binary gripper state $\hat{g}$ and next mode $\hat{m}$. As described in the main paper, the translation prediction is decomposed to first predict salient points (\cref{eq:salient}) and then predict offset $\hat{\bm{\phi}}$ w.r.t. the salient point (\cref{eq:offset}). We represent rotations in Euler angles and loss is mean squared error (MSE) between the prediction $\hat{\bm{\alpha}}$ and target $\bm{\alpha}$:
\begin{align}   
    L_\text{rot} = \| \bm{\alpha} - \hat{\bm{\alpha}} \|^2.
\end{align}
Although MSE on Euler angle ignores the wrap-around effect, i.e. $-\pi$ and $\pi$ represent the same rotation but the loss is not 0, we choose it for its simplicity and find it to work well in practice. Other representations for rotation and their corresponding losses, such as quaternion, should work similarly.
The gripper state is binary with $1$ for open and $0$ for close. 
Assuming $g$ is the ground-truth gripper state and $\hat{g}$ is the predicted probability of the gripper being open, then the gripper loss is a binary cross entropy loss:
\begin{align}
L_\text{gripper} = g \log \hat{g} + (1-g) \log (1 - \hat{g}).
\end{align}
The waypoint policy also need to predict the next mode, which is a three-way classification among candidates $\{\texttt{waypt}, \texttt{dense}, \texttt{terminate}\}$. We train it via negative log-likelihood:
\begin{align}
    L_{\text{mode}} = - \log \hat{m}
\end{align}
where $\hat{m}$ is the predicted probability for ground-truth mode $m$.
Finally, the full waypoint loss is the sum of all terms, and we find that a simple unweighted sum works well in practice, eliminating the need of additional hyperparameters:
\begin{align}
    L_\text{waypoint} = L_\text{salient} + L_\text{offset} + L_\text{rot} + L_\text{gripper} + L_\text{mode}
\end{align}

\section{Implementation Details of \algabbr}
\label{sec:impl-waypoint}

\subsection{Waypoint Policy}

The waypoint policy in \algabbr~is using a Transformer to predict salient probability and offset per point, as well as to predict rotation, gripper and mode using additional tokens similar to the \texttt{[CLS]} token in visual classification. The Transformer has $6$ layers and each layer has $512$ embedding dimensions over $8$ attention heads. We remove positional embeddings from Transformer as the point cloud input has no ordering. We set dropout to $0.1$ for all Transformer blocks to avoid overfitting. We optimize the waypoint policy with Adam~\citep{kingma2014adam} optimizer with base learning rate $1e{-}4$ and cosine learning decay over the entire training process, i.e. decaying to $0$ at the end of training. We clip the gradient with maximum norm $1$. We set batch size to 64. We also maintain an exponential moving average (EMA) of the policy with the decay rate annealing from 0 to 0.9999. We use the final EMA policy in all evaluations without any further model selection. All waypoint policies are trained for 2000 epochs.

As mentioned in the~\cref{sec:method-waypoint}, we use observations from interpolated steps as data augmentation for waypoint training, a technique we refer to as temporal augmentation:
\begin{align}
    \{(o_t, a_t, \texttt{waypt}, w_{t'}, z_{t'}) \mid  t' \leq t  \leq t' + \alpha k_{t'}\}
\end{align}
where $t'$ is the initial step when the demonstrator specified the current waypoint, $k_{t'}$ is the total number of steps it takes for the controller $\mathcal{C}$ to execute this waypoint. Essentially we train $\pi^\texttt{waypt}(w_t' | o_t)$ on the initial $o_{t'}$ as well as the intermediate observations $o_{t'+1:t'+\alpha k_{t'}}$. Here $\alpha = 0$ means no temporal augmentation and $\alpha = 1$ means training on the entire waypoint segment. 
In practice, we find that setting $\alpha=0.2$ strikes a balance between sufficient augmentation while avoiding interpolated observations that occur too late in a waypoint segment, which can cause the policy to confuse the current target waypoint with the next one. As a concrete example, the waypoint dataset for the Robomimic Square task contains 50 demonstrations, each containing 6 waypoints. The raw dataset for training the waypoint policy contains 300 examples. With the temporal augmentation, the dataset now contains roughly 1800 examples, increasing the amount of data by $6$ times.
All the waypoint policies, including vanilla waypoint baselines in~\cref{sec:waypt_exps} are trained with temporal augmentation $\alpha=0.2$ as it improves the performance for all of them. The performance of each waypoint policy with and without temporal augmentation is listed in~\cref{tab:temporal-aug}. 
We can see that temporal augmentation improves the performance of waypoint policies in five out of six scenarios and achieves similar performance in the remaining one. \algabbr~benefits the most from the augmentation technique, with $13.5\%$ average improvement across the two environments.

Apart from the temporal augmentation, we also apply random translation augmentation to the point cloud input and target action. The amount of random translation is sampled uniformly from $[-5\text{cm}, 5\text{cm}]$ along $x,y,z$-axes. We follow the common practice to crop the point cloud to remove points outside of the workspace but do not apply any other vision pipelines (i.e. no object detection nor segmentation) to preprocess the point cloud.

\begin{table}[t]
\centering
\resizebox{\columnwidth}{!}{ % Resizes the table to fit within the column width
\begin{tabular}{l c c c | c c c}
\toprule
Env & \multicolumn{3}{c|}{Can (20 demos)} & \multicolumn{3}{c}{Square (50 demos)} \\
Waypoint Method & Vanilla & Vanilla (+ Aux SP) & \algabbr & Vanilla & Vanilla (+ Aux SP) & \algabbr \\
\midrule
w/o Temp. Aug. & 68\%  & 70\% & 77.5\% & 8\%  & \textbf{68\%} & 75.5\%  \\
\cellcolor{orange!30}w/ Temp. Aug. & \cellcolor{orange!30}\textbf{70\%}  & \cellcolor{orange!30}\textbf{78\%} & \cellcolor{orange!30}\textbf{93.5\%} & \cellcolor{orange!30}\textbf{13\%} & \cellcolor{orange!30}65.5\% & \cellcolor{orange!30}\textbf{86.5\%}  \\
\bottomrule
\end{tabular}
}
\caption{Waypoint Policy Ablations}
\label{tab:temporal-aug}
\end{table}

% A benefit of our action space is that all of the intermediate poses the end-effector interpolates through to reach a particular waypoint serve as additional data points to train on, as they have the same target waypoint action and same reference salient point. We thus sample from these interpolated states during training as a cheap way to do temporal augmentation and increase the amount of observation-action pairs we train on (\cref{sec:method-waypoint}). In \cref{tab:waypoint_ablations}, we find that temporal augmentation boosts performance in nearly all cases, not only for \algabbr~ but also for the waypoint baselines. As a result, we apply temporal augmentation to train \algabbr~ in all experiments.

\subsection{Dense Policy}

The dense policy in \algabbr~is a diffusion policy. We closely follow the original implementation of~\cite{chi2023diffusion}. Specifically, we use ResNet-18~\citep{he2016deep} encoder to process the wrist image and append the proprioceptional to the image embedding before feeding it to a 1-D convolutional UNet for action denoising. The diffusion policy is trained with DDPM to predict the noise given the noisy action as input and observation as context.
We follow the best practices of training it using Adam~\citep{kingma2014adam} with weight decay, cosine learning rate schedule and take the exponential moving average of the policy as final policy for evaluation. Our implementation is able to reproduce the results on Robomimic from the original paper.

\section{Extended Related Work}

\textbf{Relationship between Salient Point and Affordance}. Affordance learning is a common concept in robotics manipulation. It can refer to classifying the nature of interaction for certain object points (e.g., a tool handle is "graspable,” a button is "pressable") ~\citep{Affordance-Learning-from-Play}, or more broad action candidates or success possibilities associated with objects~\citep{saycan2022arxiv}. Salient points in \algabbr~can be thought of as a form of per-point affordance, but must be combined with offsets to specify how the end-effector should interact with a given point. Critically, we demonstrate that the per-point affordances in \algabbr~allow the policy to focus on task-relevant features and avoid paying attention to arbitrary objects, leading to robust execution in the presence of visual distractors.

\textbf{Action Representation in Robotics.} Several works have also considered using classification instead of regression for action prediction in imitation learning setting. PerAct~\citep{shridhar2022peract} divides the entire 3D workspace into voxels and convert the end-effector pose prediction problem into classification over the fixed set of voxels. The precision of prediction depends on the granularity of the voxel, and the number of action candidates grows cubically in the number of voxels, making it challenging for tasks that requires high-precision. 
Act3D~\citep{gervet2023act3d3dfeaturefield} performs coarse-to-fine scoring for “ghost points”, which addresses the granularity issue of PerAct, but still lacks any kind of explicit intermediate representation such as salient points.
In comparison, \algabbr~first predicts the salient point, a point that \emph{physically exists} in the input point cloud, through classification and then predicts offset w.r.t. the salient to recover the full action. This allows \algabbr~ to be arbitrarily precise without incurring the high cost of having fine-grained voxels.

SGRv2~\citep{zhang2024leveraging} is a recent work that uses per-point offset prediction for robotics manipulation. It predicts actions as per-point offset for \emph{all} points, and uses a weighted average to get the final action output.
In contrast, \algabbr~utilizes \emph{salient point} learned from human labels as the anchor for offset prediction.
Due to the existence of salient points, \algabbr~applies the offset loss only to the proximal points of the “demonstrator-specified salient point”, i.e. the red points in ~\cref{fig:method-waypoint} and ~\cref{eq:offset}. This a easier task for the neural network since it does not need to allocate capacity to predict offset for points far from the salient points. 
% This also allows the network to have different position targets for different parts of the points, e.g. the offsets near the top drawer correspond to actions for opening the top drawer while the offsets near the bottom drawer correspond to actions for opening the bottom drawer. 
Additionally, the salient point learning objective is one of the main strength of \algabbr~. The ablation in the right panel of ~\cref{fig:hybrid_waypoint_results} shows that the classification loss is helpful, that even just having it as an auxiliary task significantly improves performance of the ``vanilla waypoint”.
On the implementation side, \algabbr~uses a GPT-2 style Transformer while SGRv2 uses PointNeXt as backbone. Although both methods should work well with either architecture choices but we think it is interesting to see a generic architecture not specialized for point cloud to perform quite well in these tasks.

\section{Implementation and Performance of Baselines}

In this section we discuss the steps we have taken to ensure that the performance of our baselines is valid and discuss potential reasons for the low performance for baselines like Hydra and OpenVLA.

\textbf{Diffusion policy (DP):} Our DP implementation is able to reproduce the reported results from the original paper and it achieves the reported performance on Robomimic Can and Square with 200 demonstrations from the original Robomimic dataset. It uses the same set of cameras (3rd person and wrist) as \algabbr. In the \algabbr experiment we collected 50 demonstrations for the Square and we have verified that the DP trained on the Sphinx dataset ($44\%$, 50 demonstrations) is similar to the one trained on 50 demonstrations from the Robomimic dataset ($45\%$, 50 demonstrations). 

\textbf{3D diffusion policy (DP3):} Our implementation closely follows the one from their original codebase. In their original paper, DP3 is not evaluated on the Robomimic benchmark. We find it to perform similarly as DP on the real world Drawer task but perform slightly worse than DP on Robomimic benchmarks and other more complicated real world tasks. 
This is reasonable given that DP3 purely conditions point clouds constructed from 3rd person cameras and does not use wrist camera images. The close-up information from the wrist camera images is crucial for Robomimic tasks as well as our real world tasks. In comparison, \algabbr~switches to a wrist-view image based diffusion policy for fine-manipulation, resolving the lack-of-detail issue of point cloud inputs.

\textbf{Hydra:} We modernize Hydra by using the diffusion policy as its dense policy while keep the rest of the implementation as close to the original design as possible. 
The low performance of Hydra seems unreasonable at first glance, but it can be explained via a close look at their original results and our ablations.  
Our coffee-making task is similar to the one in Hydra. However, the original Hydra paper collected 100 demonstrations with little variation on the location of the coffee machine, and put the cup and coffee pod on a shelf to make them easier to pick up. In our case, we randomize the initial location and orientation of the cup, pod and coffee machine, and only use 60 demonstrations. Therefore, it is reasonable to expect a much lower performance for Hydra on this task.
Hydra’s waypoint branch is similar to our ``vanilla waypoint", i.e. directly predicting target pose via regression, but uses images instead of point clouds. From the ablation (\cref{fig:hybrid_waypoint_results}, right panel), we see that this vanilla waypoint policy is noticeably worse than plain diffusion policy on the hard task (Square) that requires precision. Therefore, considering that Hydra’s waypoint policy is worse than diffusion policy, it is not surprising to see that the full Hydra policy (vanilla waypoint + diffusion policy) performs worse than diffusion policy in the real world tasks.

\textbf{OpenVLA:} We follow the instructions and code provided by the OpenVLA authors and apply everything as in their fine-tuning script (which does use image augmentation). We believe the poor performance of OpenVLA is not surprising for a few reasons.
From Figure 5 (left half) of the OpenVLA paper, we see that fine-tuned OpenVLA is **worse** than Diffusion Policy in 2 out of the 3 cases in single-task setting on Franka Panda hardware. In general, OpenVLA’s distribution of pre-training data relies heavily on the Bridge Dataset, which has a major embodiment gap compared to our Franka dataset.
OpenVLA is only compatible with a single 3rd person view, and cannot leverage a close-up wrist image view. On all of the real world tasks we did, the wrist camera view is crucial to sensing end-effector object alignment for precise manipulation. For example, for cup stack (the easiest of the hybrid tasks), even being off by a centimeter can lead to knocking over one or both of the cups. Thus, OpenVLA’s poor performance is unsurprising given its limited observation space.
Other recent work~\citep{pi0} reports similar trends on OpenVLA where a fine-tuned OpenVLA performs poorly, achieving $0\%$ success rate on 4 out of 5 settings and $\approx35\%$ on 1 setting.

\end{document}